# Humanoid robot pitch axis stabilization using linear quadratic regulator with fuzzy logic and capture point

Bagaskara Primastya Putra, Gabrielle Satya Mahardika, Muhammad Faris, and Adha Imam Cahyadi*

Department of Electrical and Information Engineering, Faculty of Engineering, Universitas Gadjah Mada, Yogyakarta, 55281, Indonesia
E-mail addresses: bagaskara.primastya.putra@mail.ugm.ac.id, g.satya @mail.ugm.ac.id, mfaris@ugm.ac.id, adha.imam@ugm.ac.id
*Corresponding author.

**Abstract**
This paper aims for a controller that can stabilize a position-controlled humanoid robot when standing still or walking on synthetic grass even when subjected to external disturbances. Two types of controllers are designed and implemented: ankle strategy and stepping strategy. The robot's joints consist of position-controlled servos which can be complicated to model analytically due to nonlinearities and non-measurable parameters, hence the dynamic model of the humanoid robot is acquired using a non-recursive least squares system identification. This model is also used to design a Kalman Filter to estimate the system states from the noisy inertial measurement unit (IMU) sensor and design a linear quadratic regulator (LQR) controller. To handle the nonlinearities, the LQR controller is extended with a fuzzy logic algorithm that changes the LQR gain value based on angle and angular velocity membership functions. The proposed control system can maintain the humanoid robot's stability around the pitch axis when subject to pendulum disturbances or even restraining force from a spring balance.

## 1 Introduction

Humanoid robots are designed to mimic human motion to accomplish tasks that requires human capabilities or even aid in tasks that couldn't be accomplished alone by humans. Applications of humanoid robots include cargo lifting [1], rescue missions [2], soccer-playing [3], etc. However, humanoid robots are prone to falling as they have a high center of mass (CoM) and a small supporting area on their feet. This gets even worse when the robot is subject to disturbances such as uneven terrain, collisions with obstacles and other robots, etc.

The majority of commercially available humanoid robots are position-controlled such as ROBOTIS OP3 [4] and NAO [5]. This means the servo joint actuators are controlled by commanding reference angular position. This configuration has the advantage of easier pattern generation and trajectory control by using inverse kinematics. But for disturbance rejection, the dynamics of the humanoid robot must be accounted for. The first approach is joint torque calculation which requires the full-body model of the robot and force/torque sensors [6],[7],[8],[9]. This approach tends to be expensive. Another approach is by using a simplified model of the robot to generate fast push recovery behaviors [10],[11],[12]. This approach is more suitable for our kid-size humanoid robot intended for playing soccer on a synthetic grass environment.

Thus, this work aims to develop a disturbance rejection control approach that can stabilize a position-controlled humanoid robot without the use of force/torque sensors. It must be able to stabilize the robot when standing still, walking on synthetic grass, and even when subject to external disturbances.

## 2 Related work

Numerous works have been done by researchers to stabilize position-controlled humanoid robots without the use of force/torque sensors. A push recovery controller based on capture point feedback control has been implemented [12]. A proportional derivative (PD) controller is implemented to apply control effort on the servo joint angles for the ankle and hip strategy. This controller has the advantage of being simple yet able to reject impulsive disturbances using only the ankle and hip when stepping is not allowed. However, the classical PD controller is difficult to optimize based on environmental constraints. A PD controller has also been implemented to control the posture and gait with feedback from the CoM position [13]. This controller enables the robot to walk on inclined surfaces.

A push recovery controller using the ankle and stepping strategy has been implemented [14]. The ankle strategy uses an LQR controller to generate the required ankle torque based on LIPM, then converted into ankle joint angle. The stepping strategy uses the capture point method to stop the robot from falling when the ankle strategy is incapable of handling the disturbance. However, this controller can only reject impulsive disturbances coming from behind the robot. After the stepping strategy has been activated, the robot remains in the current pose unable to recover to its standing pose.

Fuzzy logic approaches have also been implemented. A joint controller using a fuzzy PD algorithm with a single neuron self-adapting PID has been implemented [15]. This configuration can improve the response speed, control precision, and robustness of the robot's joint. However, disturbances on the whole humanoid robot have yet to be tested. A comparison of the classical PI+D controller to the fuzzy modified PI+D controller has been analyzed on a simulated mathematical model of the robot [16]. The fuzzy PI+D controller can improve performance, especially when dealing with unexpected disturbances. A fuzzy PD controller has been implemented on an NAO robot to track a reference trajectory [17]. The robot can still walk while following the reference trajectory even when subject to external disturbances. Fuzzy logic control has also been implemented to stabilize a robot on top of an unstable bongo board platform [18]. In terms of performance, the fuzzy logic control is equally capable compared to PID control. The PID control can automatically recover from sudden disturbances but results in larger oscillations. However, the fuzzy logic control can maintain a longer stable position with the lowest deck inclination but was unable to automatically recover from sudden disturbances.

Reinforcement learning (RL) has been implemented as a push recovery controller which consists of three main strategies: ankle, hip, and stepping strategy [11]. RL is applied to map the robot walking state to the best of these three strategies for it to be able to reject external disturbances. The training process is done on a servo-controlled platform to simulate impulsive disturbance while commanding the robot to walk on it in different directions. This push recovery controller has the advantage of learning directly on physical hardware. However, the training setup seems complex as it requires a mechanized moving platform.

A linear inverted pendulum model (LIPM)-based balance controller has been implemented for an open-loop omnidirectional central pattern generated walking gait [19]. This capture step framework can reject disturbances such as pushes, collisions, and opponent's feet by modifying the footstep location as well as its timing. An online learning balance controller [20] can complement and improve the previously mentioned analytic footstep controller.

## 3 Contributions

The main contributions of this work are as follows:
- Design and implementation of a disturbance rejection controller for a position-controlled humanoid robot. This controller consists of an ankle strategy and a stepping strategy.
- System identification of a position-controlled humanoid robot using non-recursive least squares method.
- Ankle strategy for controlling the robot's CoM by applying control effort at the ankle pitch joints. This strategy consists of a state feedback controller with LQR gains extended with fuzzy logic and Kalman Filter to estimate states from noisy sensors. Nonlinearities are still taken into account while still being computationally efficient.
- Stepping strategy by commanding the robot's walking amplitude to place its foot in the estimated capture point. This complements the ankle strategy when it is no longer able to reject large and constant disturbances.

The implemented control algorithm is shown in **Fig. 1**. The algorithm consists of two main sections which are sensor data processing and the main controller. The walking gait can be substituted with any open-loop walking gait as long as it is stable when not subject to disturbances. We implemented the QuinticWalk walking gait from Hamburg Bit-Bots [21], [22] which is based on Rhoban FC [23].

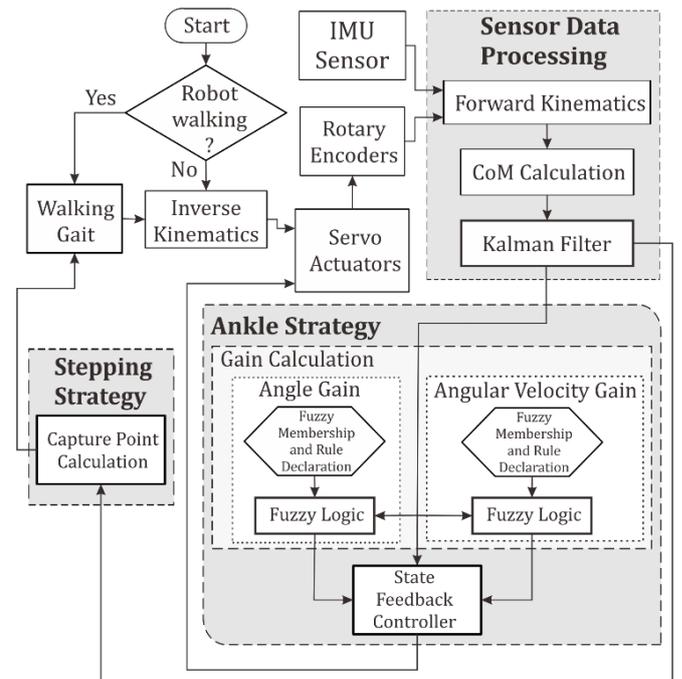

**Fig. 1.** Control algorithm implemented in our robot

## 4 Robot configuration and modeling

Our robot is a kid-size humanoid robot with 20 degrees of freedom (DoF) as shown in **Fig. 2**. The upper torso has 8 DoF which consists of a 2 DoF head and a pair of 3 DoF arms. While a pair of 6 DoF legs construct the lower body. Two types of servo actuators are installed which are Dynamixel MX-28 for the upper torso and Dynamixel MX-64 for the legs. Dynamixel MX-28 is chosen for the arms and head as they require less torque. While the legs that support the whole body require more torque so the Dynamixel MX-64 is used.

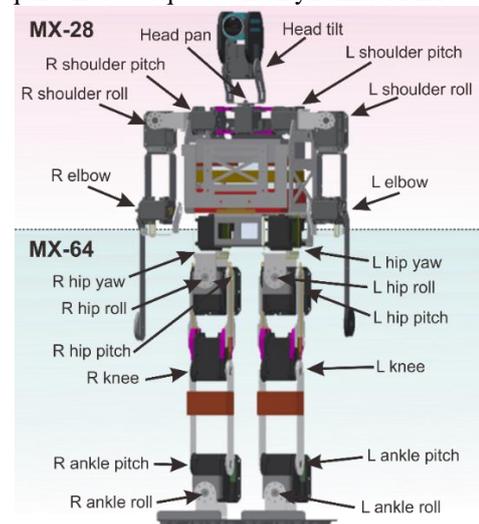

**Fig. 2.** Robot mechanical design and servo configuration

A humanoid robot is a nonlinear and unstable system due to its high center of mass and small supporting feet area. One of the humanoid robot stability criteria is the zero moment point (ZMP). ZMP is the point where the moment of the ground reaction force equals zero because it is perpendicular to the ground plane [24]. It is also called the center of pressure (CoP). For a humanoid robot to maintain its stability, the ZMP must be kept inside the support polygon. Support polygon is the convex hull of the area of an object that is in direct contact with the ground [25]. The ZMP can be measured using force sensing registers (FSR) installed under the robot's feet as shown in **Fig. 3**.

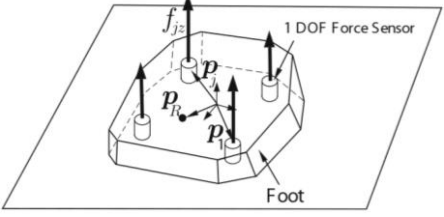

**Fig. 3.** Calculation of ZMP by multiple force sensors [25]

In our case, the robot does not have force/torque sensors. This means that the ZMP cannot be directly measured. The ZMP in *x* and *y*-axis ($p_x$ and $p_y$) can also be approximated by using the following equations:

$$p_x = x - \frac{(z - p_z)\ddot{x}}{\ddot{z} + g} \quad (1)$$

$$p_y = y - \frac{(z - p_z)\ddot{y}}{\ddot{z} + g} \quad (2)$$

where:
- $c = [x \quad y \quad z]$ is the robot's center of mass (CoM)
- $\ddot{c} = [\ddot{x} \quad \ddot{y} \quad \ddot{z}]^T$ is the linear acceleration of the robot's CoM

This approximation method is also not applicable in our humanoid robot as the acceleration data from our IMU sensor is very noisy.

For the above reasons, ZMP-based control is not applicable for our humanoid robot. However, we implemented a simpler stability criterion which is based on the CoM angle and estimated capture point using measurements from servo joint rotary encoders and IMU sensor.

The control algorithm shown in **Fig. 1** is implemented in an Intel NUC Mini PC. The software is developed using Ubuntu 16.04 operating system and Robot Operating System (ROS) as the main framework.

## 5 System identification

Modeling a position-controlled humanoid robot system analytically from a linear inverted pendulum model (LIPM) is a complicated task. The control effort of LIPM is torque. While in the case of a position-controlled humanoid robot, the control effort is the reference angular position of the servo. The more complex whole system dynamics of the actual robot cannot be represented using LIPM. This complexity is caused by the inability to directly measure the robot's moment of inertia and uncertainty that hasn't been considered such as servo backlash and friction. This robot is also required to walk on synthetic grass which further increases the system's complexity.

To acquire a better model of the humanoid robot, a system identification approach can be utilized. We implemented the non-recursive least squares method [26] as it is easy to acquire a simple model with the number of states equal to the number of outputs and fairly accurate model validation.

This method aims to minimize the square error between the model output and the measured output. The state space parameters can be estimated from the input and output time series. The output time series can be written into Equation (3).

$$Y_k = \Psi_k \widehat{\Theta}_k + E_k \quad (3)$$

where:
- $Y_k$ is the output time series
- $\Psi_k$ is the identification data matrix consisting of state and input time series
- $\widehat{\Theta}_k$ is the estimated state space parameter matrix
- $E_k$ is the error between measured and model output

The state space parameters can be solved by minimizing the square of the error $E_k$ which results in Equation (4).

$$\widehat{\Theta}_k = (\Psi_k^T \Psi_k)^{-1} \Psi_k^T Y_k \quad (4)$$

The states which are also the outputs of this humanoid robot system are $\theta$ (the angle between the robot's CoM and vertical plane) and $\dot{\theta}$ (angular velocity of the robot's CoM) around the pitch or y-axis. While the input to this system is the ankle pitch servo angle $\theta_{anklePitch}$. This is shown clearly in **Fig. 4**.

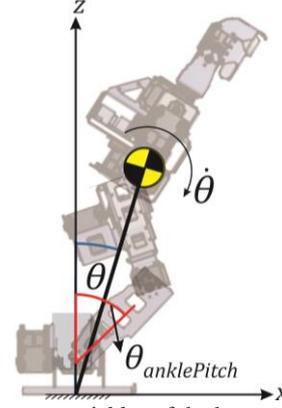

**Fig. 4.** Input and output variables of the humanoid robot model

The system identification process that is implemented in our humanoid robot is shown in **Fig. 5**.

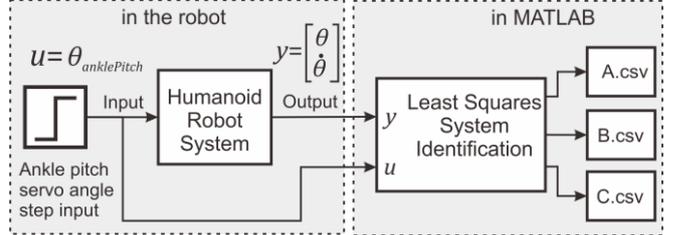

**Fig. 5.** System identification implementation

The step response of the actual humanoid robot system is acquired by applying a step input to the ankle pitch servo angle. Then the recorded input and output time series are processed using the non-recursive least squares system identification (Equation (4)) implemented in a MATLAB script. This script produces the discrete-time state space parameters shown in **Table 1**.

**Table 1.** Identified state space parameters

| $A = \begin{bmatrix} 0.995 & 0.021 \\ -0.584 & 0.879 \end{bmatrix}$ | $B = \begin{bmatrix} 0.013 \\ 1.416 \end{bmatrix}$ |
|---|---|
| $C = \begin{bmatrix} 1 & 0 \\ 0 & 1 \end{bmatrix}$ | Sampling frequency: 41.664 Hz |

This identified model is validated by comparing the identified model's step response with the real system's step response as shown in **Fig. 6**.

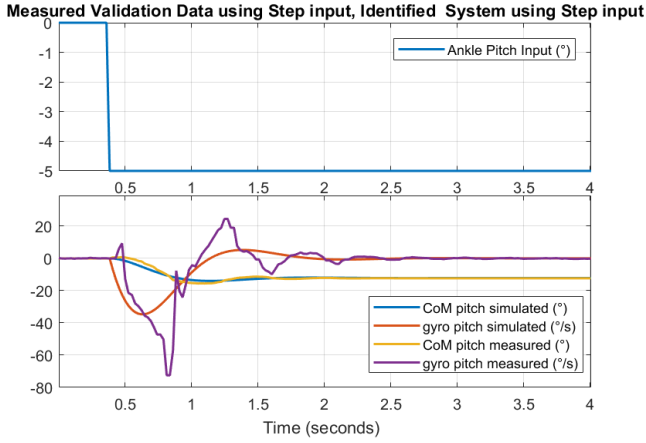

**Fig. 6.** Step response model validation

The model validation can be measured using variance accounted for (VAF) [27] shown in Equation (5). VAF shows the percentage of the variance of the difference between the measured output $y_k$ and estimated output $\hat{y}_k$ to the variance of measured output $y_k$. The higher the VAF, the more accurate the model.

$$VAF(y_k, \hat{y}_k) = \max\left(0, \left(1 - \frac{\frac{1}{N}\sum_{k=1}^{N}\|y_k - \hat{y}_k\|_2^2}{\frac{1}{N}\sum_{k=1}^{N}\|y_k\|_2^2}\right) \cdot 100\%\right) \quad (5)$$

The VAF of the identified model is shown in **Table 2**. The angle output VAF has an accuracy close to 100%, while the angular velocity output VAF is not as accurate. This is caused by the identified model only having an order of two. In the real humanoid robot system, the system order can be more than two and nonlinear. Nonetheless, the simulated model angular velocity response can still track the measured angular velocity response, as shown in **Fig. 6**. The simulated model response is like a filtered version of the measured response.

**Table 2.** Identified model VAF

| Angle output VAF | Angular velocity output VAF |
|---|---|
| 96.042% | 61.485% |

## 6 Ankle strategy

The ankle strategy is a strategy of stabilizing the robot's body by manipulating the ankle pitch joint angle. This strategy consists of a Kalman Filtered state feedback controller with LQR gain extended with fuzzy logic shown in **Fig. 7**.

### 6.1 Model-based control

#### 6.1.1 Kalman filter

Our robot's IMU sensor has an accurate orientation measurement, but noisy gyro measurement. If this noisy measurement is directly used for feedback control, it would cause oscillations and instability. To solve this problem, we implemented the Kalman Filter. The discrete-time state equation of the Kalman Filter is shown in Equation (6) and the right section of **Fig. 7**.

$$\hat{x}_{k+1} = A\hat{x}_k + Bu_k + K_f(y_k - C\hat{x}_k) \quad (6)$$

The Kalman Gain can be solved using Equation (7)

$$K_f = APC^T(V_n + CPC^T) \quad (7)$$

where $P$ is the Riccati variable that satisfies the algebraic Riccati equation (ARE) shown in Equation (8).

$$P = APA^T + V_d - APC^T(V_n + CPC^T)^{-1}CPA^T \quad (8)$$

The identified model from **Table 1** can be used as the state space model for the Kalman Filter. While the covariance matrices $V_d$ and $V_n$ need to be tuned in a way that the estimated states trust the sensors more than the model or trust the model more than the sensors. The implemented matrices in our robot are shown in **Table 3**.

**Table 3.** Process noise covariance matrix $V_d$ and measurement noise covariance matrix $V_n$

| $V_d = \begin{bmatrix} 1 & 0 \\ 0 & 1 \end{bmatrix}$ | $V_n = \begin{bmatrix} 10^{-6} & 0 \\ 0 & V_{n2,2} \end{bmatrix}$ |
|---|---|

The first row and column element of matrices $V_d$ and $V_n$ is given those values because the angular position data from IMU is reliable enough. This means the first element of matrix $V_n$ is set to a small as possible value because the angular position data has very little noise.

On the other hand, the angular velocity data from the gyro sensor in the IMU is very noisy. This means the second row and column element of matrix $V_d$ and $V_n$ needs to be tuned. For ease of tuning, matrix $V_d$ is set to a 2×2 identity matrix so that we only tune the values in matrix $V_n$. This means the value of $V_{n2,2}$ in matrix $V_n$ is tuned to compensate for the noise.

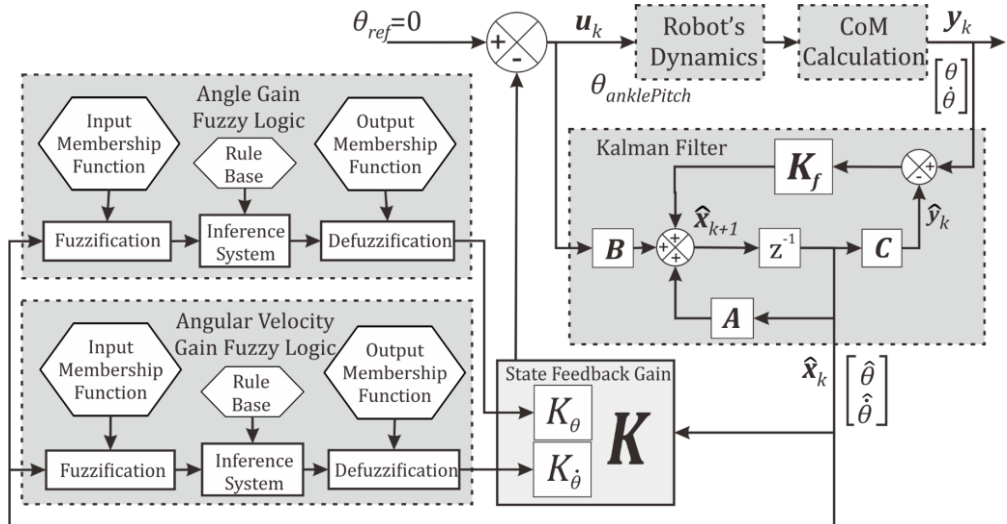

**Fig. 7.** Ankle strategy control block diagram

### 6.1.2 Linear quadratic regulator

In chapter 5, a discrete-time state space model of the humanoid robot is acquired. In theory, a model-based control approach can be implemented to control the states of this system. One of the model-based controllers is a linear quadratic regulator (LQR) and in this case a discrete-time LQR. It is a state feedback optimal controller that minimizes an objective scalar cost function or performance index in discrete-time [28]. The cost function that is minimized is shown in Equation (9).

$$J = \frac{1}{2}\sum_{k=0}^{\infty} x_k^T Q x_k + u_k^T R u_k \quad (9)$$

where:
- $Q$ is the state cost matrix
- $R$ is the input cost matrix

The objective of this controller is to calculate the control effort $u_k$ that minimizes the cost function from Equation (9)

$$u_k = -K x_k \quad (10)$$

in which the state feedback gain $K$ can be calculated using Equation (11).

$$K = (B^T P B + R)^{-1}(B^T P A) \quad (11)$$

The Riccati variable $P$ can be calculated by solving the ARE in Equation (12).

$$A^T P A - P - (A^T P B)(B^T P B + R)^{-1}(B^T P A) + Q = 0 \quad (12)$$

The identified model from **Table 1** can be used as the state space model for the LQR. While the cost matrices $Q$ and $R$ must be tuned to produce the desired closed-loop response.

**Table 4.** State cost matrix $Q$ and input cost matrix $R$

| $Q = \begin{bmatrix} Q_{1,1} & 0 \\ 0 & 1 \end{bmatrix}$ | $R = 1$ |
|---|---|

For ease of tuning, only specific matrix elements are tuned shown in **Table 4**. Matrix $R$ is set to identity or one because the system only has one input. The second row and column element of matrix $Q$ which contributes to the angular velocity response is also set to one. The reason is that the output that we want to control is the CoM angle of the robot. Thus, only the value of $Q_{1,1}$ in matrix $Q$ is tuned in order to produce the desired closed-loop response.

### 6.2 Fuzzy logic

#### 6.2.1 System architecture

To increase the robustness of the LQR controller, fuzzy logic is implemented to calculate the state feedback gain. A rule-based fuzzy system with a singleton model is implemented to give ease in mapping the input and output relations. The rule-based fuzzy system is used to map the relations between the inputs and outputs.

The overall state feedback gain calculation process is shown in the left section of **Fig. 7**. The model-based control explained in chapter 6.1 contains two states which are CoM angle $\theta$ and angular velocity $\dot{\theta}$. Because of this, two fuzzy systems are implemented in parallel to calculate each state feedback gain ($K_\theta$ and $K_{\dot{\theta}}$).

#### 6.2.2 Fuzzy membership function and rule declaration

The fuzzy membership function and rule must be declared before executing the fuzzy process. Each fuzzy system consists of two same inputs which are the CoM angle and angular velocity with the same input and output membership function, so in total there are four fuzzy membership functions: (1) angle input membership function, (2) angular velocity input membership function, (3) angle gain output membership function, and (4) angular velocity gain output membership function.

These membership functions are designed with the basic shape of a trapezoid. The structure of this membership function is shown in **Fig. 8**.

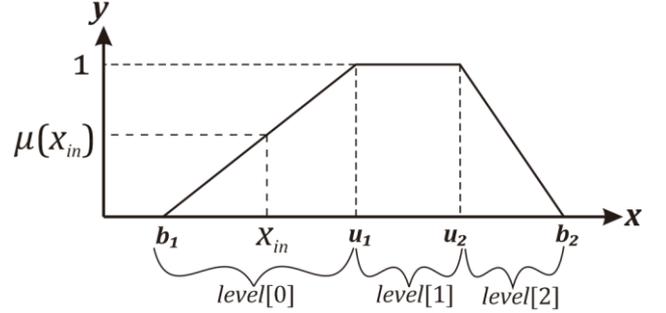

**Fig. 8.** Trapezoidal membership function general structure

where:
- $x_{in}$ is the crisp input of the membership function in the $x$-axis,
- $\mu(x_{in})$ is the truth value of the membership function in the $y$-axis for its corresponding $x_{in}$ input,
- $b_1$ is the $x$-axis coordinate of the trapezoid's bottom left corner,
- $b_2$ is the $x$-axis coordinate of the trapezoid's bottom right corner,
- $u_1$ is the $x$-axis coordinate of the trapezoid's upper left corner,
- $u_2$ is the $x$-axis coordinate of the trapezoid's upper right corner,
- $level[0]$ is the trapezoid section or boundary of the membership function with a positive slope,
- $level[1]$ is the horizontal roof section of the trapezoid or the core of the membership function, and
- $level[2]$ is the trapezoid section or boundary of the membership function with a negative slope.

The membership function can also be the shape of a triangle. If this is the case, then the trapezoid can be simplified with $u_1 = u_2$ and the $level[1]$ section can be neglected.

The fuzzy rules can be represented in a tabular form with its output fuzzy sets corresponding to its row and column input fuzzy sets shown in **Table 5**.

**Table 5.** Fuzzy rule table

| Angular velocity \ Angle | Negative (N) | Average (A) | Positive (P) |
|---|---|---|---|
| **Negative High (NH)** | Output [0][0] | Output [0][1] | Output [0][2] |
| **Negative (N)** | Output [1][0] | Output [1][1] | Output [1][2] |
| **Positive (P)** | Output [2][0] | Output [2][1] | Output [2][2] |
| **Positive High (PH)** | Output [3][0] | Output [3][1] | Output [3][2] |

The values of these membership functions and rules are later tuned through experiments using various angle state cost values $Q_{1,1}$ of the LQR gain. The tuned fuzzy rule table values are later shown in **Table 14** and **Table 15**.

### 6.2.3 Fuzzification

The fuzzy process starts with fuzzification. It is a process of converting crisp input data into fuzzy sets using membership functions [29]. The fuzzification process will produce the fuzzy sets that will later be the input of the inference system.

The implemented fuzzification algorithm is shown in **Algorithm 1**. The condition of the crisp input $x_{in}$ is first checked based on the trapezoid *level* sections mentioned in **Fig. 8** before calculating its corresponding fuzzy output $\mu(x_{in})$.

**Algorithm 1.** Fuzzification algorithm

1:    **function** FUZZIFICATION($x_{in}$)
2:       **for** $i \leftarrow 0$ **to** $N-1$ **do**
3:          **if** $b_1[i] \leq x_{in} < u_1[i]$ **then**
4:             $level[0] = i$
5:          **end if**
6:          **if** $u_1[i] \leq x_{in} < u_2[i]$ **then**
7:             $level[1] = i$
8:          **end if**
9:          **if** $u_2[i] \leq x_{in} < b_2[i]$ **then**
10:            $level[2] = i$
11:          **end if**
12:      **end for**
13:      **for** $i \leftarrow 0$ **to** $N-1$ **do**
14:         **if** $level[0] = i$ **then**
15:            $y_{fuzz}[level[0]] = \mu(x_{in}, b_1[i-1], u_1[i-1], 0, 1)$
16:         **end if**
17:         **if** $level[1] = i$ **then**
18:            $y_{fuzz}[level[1]] = 1$
19:         **end if**
20:         **if** $level[2] = i$ **then**
21:            $y_{fuzz}[level[2]] = \mu(x_{in}, u_2[i-1], b_2[i-1], 1, 0)$
22:         **end if**
23:      **end for**
24:      **return** $y_{fuzz}$
25:    **end function**

where:
- $N$ is the number of input membership functions,
- $y_{fuzz}[i]$ is the output fuzzy value of the $i$-th membership function, and
- $\mu(x_{in}, x_1, x_2, y_1, y_2) = \frac{y_2 - y_1}{x_2 - x_1} \cdot (x_{in} - x_1) + y_1$ is the input membership function.

### 6.2.4 Inference system

The next step of the fuzzy logic algorithm is the inference system. The inference system is the relation between input and output variables using defined fuzzy rules to produce fuzzy outputs based on membership functions [29]. In this step, the membership of the fuzzy set from the fuzzification process will be calculated.

The max-min (Mamdani) inference system is implemented as shown in **Algorithm 2**.

**Algorithm 2.** Inference system algorithm

1:    **function** INFERENCESYSTEM($y_{fuzz1}, y_{fuzz2}, N_1, N_2$)
2:       $y_{inf} = \mathbf{0} \in \mathbb{R}^{1 \times N_y}$
3:       **for** $i \leftarrow 0$ **to** $N_1 - 1$ **do**
4:          **for** $j \leftarrow 0$ **to** $N_2 - 1$ **do**
5:             $k = rule[i][j] - 1$
6:             $y_{inf}[k] = \max(y_{inf}[k], \min(y_{fuzz1}[i], y_{fuzz2}[j]))$
7:          **end for**
8:       **end for**
9:       **return** $y_{inf}$
10:    **end function**

where:
- $N_y$ is the number of output membership functions,
- $y_{fuzz1}$ and $y_{fuzz2}$ are the fuzzy sets from the fuzzification process,
- $N_1$ and $N_2$ are the number of input membership functions, and
- $rule[i][j]$ refers to the fuzzy rule mentioned in **Table 5**.

For our application of state feedback gain calculation, $y_{fuzz1}$ and $y_{fuzz2}$ are substituted with the CoM angle and angular velocity fuzzy sets respectively, while $N_1$ and $N_2$ are substituted with the number of input membership functions of the CoM angle and angular velocity respectively.

### 6.2.5 Defuzzification

After acquiring the membership result from the inference system, the next part of the fuzzy process is defuzzification. It is a process of converting fuzzy sets into crisp output using output membership functions [29].

The centroid/CoM method shown in Equation (13) is applied in this defuzzification process.

$$CoM_x = \frac{\int x\mu(x)\,dx}{\int \mu(x)\,dx} \quad (13)$$

The membership functions can intersect each other as illustrated in **Fig. 9**. Therefore, the centroid of the union of shapes A and B must be calculated. The union's area is the sum of the individual's area minus the intersection area.

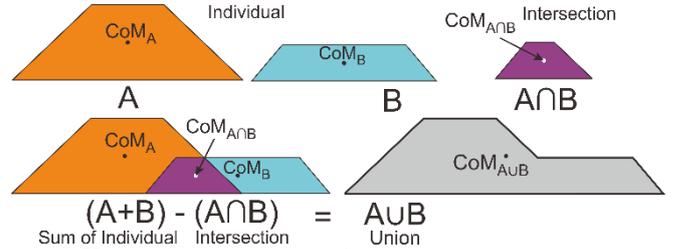

**Fig. 9.** Union of two intersecting shapes

The union's area principle can be applied to derive the algorithm for calculating the centroid of multiple membership functions. This principle can be applied to derive Equation (13) into **Algorithm 3**.

**Algorithm 3.** Centroid defuzzification algorithm

1:    **function** DEFUZZIFICATION($y_{inf}$)
2:       **for** $i \leftarrow 0$ **to** $N-1$ **do**
3:          $[Area, CoM_x] = $ CALCINDIVIDUAL($i, y_{inf}[i]$)
4:          $numerator \mathrel{+}= Area \cdot CoM_x$
5:          $denominator \mathrel{+}= Area$

```
6:      end for
7:      for j ← 0 to N − 2 do
8:          [Area, CoM_x] =    CALCINTERSECTION
            (j, j + 1, y_inf[j], y_inf[j + 1])
9:          numerator −= Area · CoM_x
10:         denominator −= Area
11:     end for
12:     return CoM_{xUnion} = numerator/denominator
13: end function
```

where $N$ is the number of output membership functions.

The CALCINDIVIDUAL() and CALCINTERSECTION() functions called in **Algorithm 3** are explained in (1) Individual calculation and (2) Intersection calculation respectively.

(1) Individual calculation

The individual membership area and centroid calculation can be achieved by first checking the condition of $y_{inf}$ from the inference system as shown in **Fig. 10**. Based on the value of $y_{inf}$, the membership's shape can either be a triangle, trapezoid, or none at all. A further detailed explanation of the area and centroid calculation process is elaborated in Appendix A.

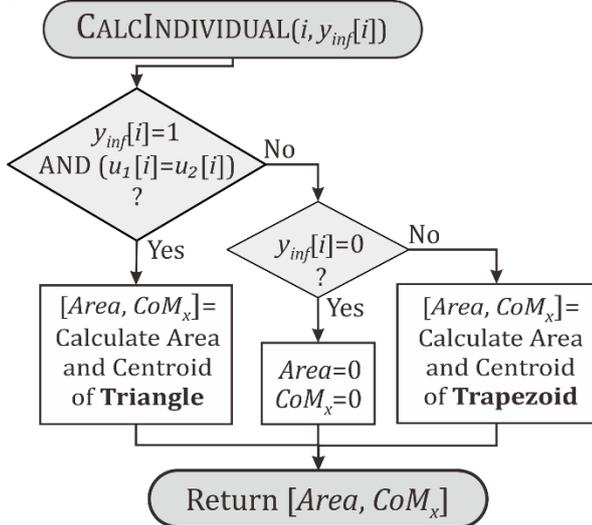

**Fig. 10.** Individual area and centroid calculation

(2) Intersection calculation

After calculating the individual memberships' area and centroid, the next step is calculating the intersection. The process of area and centroid calculation of the intersection is shown in **Fig. 11**. The membership's shape can either be a triangle, trapezoid, or none at all based on the values of $y_{inf}[j]$, $y_{inf}[j + 1]$, and the intersection peak's height. A further detailed explanation of the area and centroid calculation process is elaborated in Appendix B.

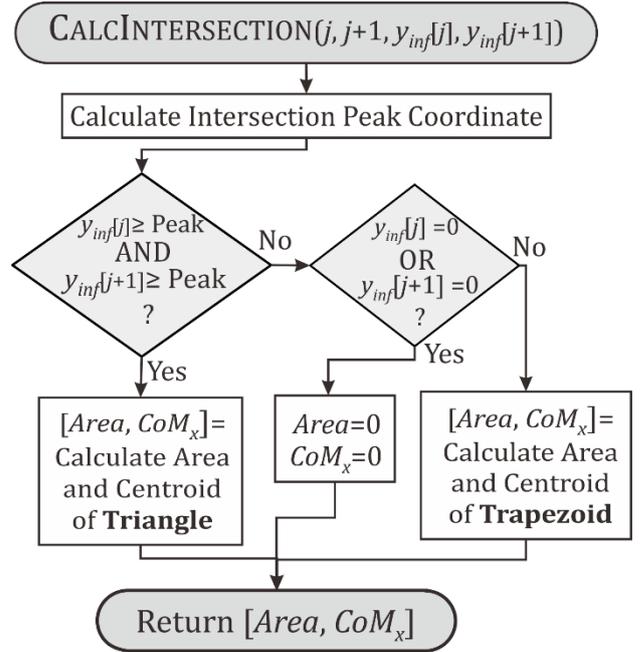

**Fig. 11.** Intersection area and centroid calculation

## 7 Stepping strategy

The stepping strategy is a humanoid robot stabilization strategy that is accomplished by placing its foot onto a point on the ground. This method is implemented because of the small support polygon of the humanoid robot causing the position control to be very limited in the equilibrium region. This stepping strategy is implemented by calculating the capture point.

### 7.1    Capture point

Capture point (CP) is the region on the ground where the robot must step to in order to bring itself to a complete stop [10]. This CP can be calculated by multiplying the CoM linear velocity with the inverted pendulum period shown in Equation (14). In [30], this CP is the combination of the robot's CoM position, its linear and angular velocity, and some offset value shown in Equation (15).

$$x_{CP} = \dot{x}_{CoM} \sqrt{\frac{z_{CoM}}{g}} \quad (14)$$

$$x_{CP} = x_{CoM} + \frac{\dot{x}_{CoM}}{\omega} + x_{offset} \quad (15)$$

But in reality, the CoM linear velocity $\dot{x}_{CoM}$ cannot be directly measured. Because of that, by utilizing the available gyro (angular velocity) sensor, the CoM linear velocity can be approximated using Equation (16) that shows the relation between the angular and linear velocity as shown in **Fig. 12**.

$$\dot{x}_{CoM} = \dot{\theta} z_{CoM} \quad (16)$$

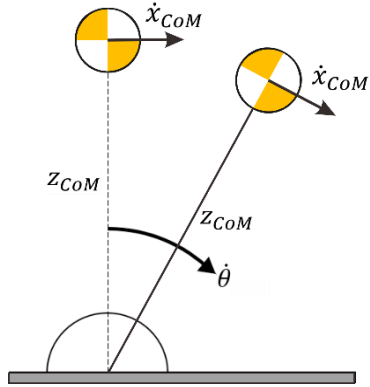

**Fig. 12.** Linear velocity and angular velocity relation

Equations (14)-(16) can be combined to form our implemented stepping strategy as shown in Equation (17).

$$x_{CP} = \dot{\theta} z_{CoM} \sqrt{\frac{z_{CoM}}{g}} + x_{offset} \quad (17)$$

The raw gyro data is very noisy so using it directly for the stepping strategy is not recommended. For this reason, the estimated angular velocity state from Kalman Filter $\hat{\dot{\theta}}$ (chapter 6.1.1) is implemented in the CP calculation. However, this Kalman Filter state estimation tends to drift when the CoM angle is large. Because of this, an algorithm is developed to dampen this angular velocity estimation to zero when its value is relatively constant in a certain amount of time. This is achieved by comparing the current estimated angular velocity with 10% of the previous estimated angular velocity. The implemented CP calculation algorithm is shown in **Algorithm 4**.

| **Algorithm 4.** Implemented capture point calculation algorithm ||
|---|---|
| 1: | **function** CALCULATECAPTUREPOINT() |
| 2: | **if** $\left\| \left( \hat{\dot{\theta}} - \dot{\theta}_{last} \right) / \dot{\theta}_{last} \right\| < 0.1$ **then** |
| 3: | $counterGyro\mathrel{+}= 1$ |
| 4: | **else** |
| 5: | $counterGyro = 0$ |
| 6: | **end if** |
| 7: | $\dot{\theta}_{CP} = \hat{\dot{\theta}} e^{-counterGyro}$ |
| 8: | $\dot{\theta}_{last} = \hat{\dot{\theta}}$ |
| 9: | $x_{CP} = \dot{\theta}_{CP} z_{CoM} \sqrt{\frac{z_{CoM}}{g}} + x_{offset}$ |
| 10: | **return** $x_{CP}$ |
| 11: | **end function** |

## 8 Experiments and results

### 8.1 Ankle strategy

#### 8.1.1 Experimental setup

The ankle strategy is tested by swinging a pendulum to the robot's torso when the robot is standing still. The pendulum is released from a stationary amplitude angle and then hits the robot's torso at its maximum velocity. In **Fig. 13**, the experimental setup for the ankle strategy is shown. A supporting metal frame is used to hang the water bottle pendulum with the string attached to the horizontal beam. The water bottle has a total mass of 760 grams. The length of the string until the center of mass of the water bottle is 1 meter. In the center of the horizontal beam, a protractor is installed perpendicular to the beam to measure the pendulum amplitude around the vertical plane. A webcam and laptop setup is placed beside it to record the robot's response.

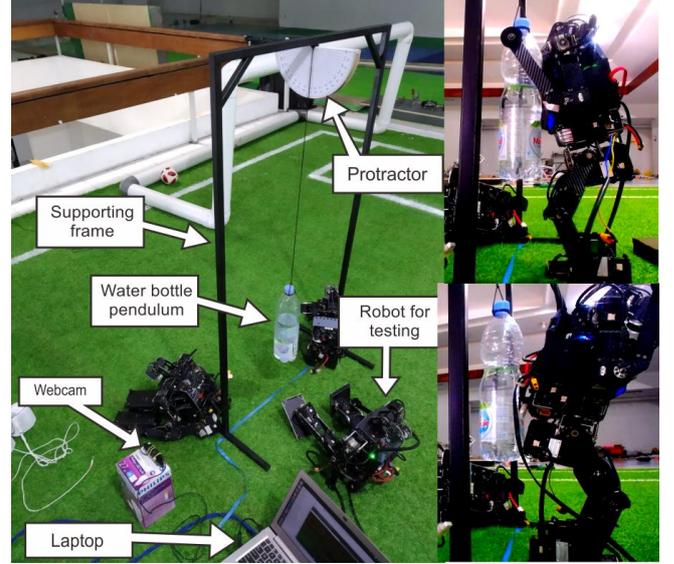

**Fig. 13.** Ankle strategy experimental setup

#### 8.1.2 Model-based control

##### 1.1.1.1 Tuning and results analysis

Before testing with the fuzzy configuration, the value of $Q_{1,1}$ from chapter 6.1.2 and $V_{n2,2}$ from chapter 6.1.1 must be tuned. The ankle strategy testing results with various tuning configurations are shown in **Table 6**. The checkmarks indicate the maximum pendulum amplitude angle that the robot can handle before falling. The black patches indicate for that amplitude angle the robot is guaranteed to fall.

**Table 6.** $Q_{1,1}$ and $V_{n2,2}$ tuning results for the ankle strategy

| No. | Disturbance origin | Configuration || Pendulum amplitude angle (degrees) ||||||
|---|---|---|---|---|---|---|---|---|---|
| | | | | 22 | 23 | 24 | 25 | 26 | 27 |
| 1 | In front of the robot | No feedback || | | | ✓ | | |
| | | $Q_{1,1}$ | $V_{n2,2}$ | | | | | | |
| 2 | | 30 | 20 | | ✓ | | | | |
| 3 | | | 35 | | | | ✓ | | |
| 4 | | | 50 | | | | ✓ | | |
| 5 | | 40 | 20 | | | | ✓ | | |
| 6 | | | 35 | | | | | ✓ | |
| 7 | | | 50 | | ✓ | | | | |
| 8 | | 50 | 20 | | | | ✓ | | |
| 9 | | | 35 | ✓ | | | | | |
| 10 | | | 50 | | | | ✓ | | |
| 11 | Behind the robot | No feedback || | | | | ✓ | |
| | | $Q_{1,1}$ | $V_{n2,2}$ | | | | | | |
| 12 | | 30 | 20 | | | | ✓ | | |
| 13 | | | 35 | | | ✓ | | | |
| 14 | | | 50 | ✓ | | | | | |
| 15 | | 40 | 20 | | | | ✓ | | |
| 16 | | | 35 | | | | ✓ | | |
| 17 | | | 50 | | | | ✓ | | |
| 18 | | 50 | 20 | | | ✓ | | | |
| 19 | | | 35 | | | | | ✓ | |
| 20 | | | 50 | | | ✓ | | | |

When the pendulum is swung from the front side of the robot, the configuration with $Q_{1,1} = 40$ and $V_{n2,2} = 35$ could handle the maximum pendulum amplitude of 26 degrees. This configuration is also better than without any feedback at all that can only handle 25 degrees.

In the case of the pendulum swung from behind the robot, there are actually four configurations that can handle the maximum pendulum amplitude of 25 degrees. However, even these configurations are no better than the no feedback configuration which can handle 26 degrees. This is caused by the forward-leaning posture of the robot shown in the right section of **Fig. 13**. The $Q_{1,1} = 40$ and $V_{n2,2} = 35$ configuration, that was the best configuration for disturbance coming from the front, could only handle 24 degrees for disturbance coming from behind. For the sake of simplicity, only one configuration is chosen. The $Q_{1,1} = 40$ and $V_{n2,2} = 35$ configuration is chosen as the best configuration.

The larger the value of $Q_{1,1}$, the LQR controller prioritizes the angle state performance. This causes the gain $K$ value & the control effort to increase. Thus, the transient response becomes faster. However, with an increasing value of $K$, there is a chance of overshoot that may cause the robot to fall. Because of that, the value of $Q_{1,1}$ must be tuned properly.

The larger the value of $V_{n2,2}$, the Kalman Filter perceives this as a larger measurement noise covariance. This causes the Kalman Filter to trust the state estimate from the model more than the measurement. This means a larger value of $V_{n2,2}$ causes the state estimate to be less sensitive to changes in the measurement. In the case of the robot given some disturbance, a less sensitive state estimate may cause the control effort to lag when countering the disturbance. However if the $V_{n2,2}$ value is too small, the system becomes oversensitive to the measurement noise. This may also cause the system to overshoot and the robot could fall. For this reason, the combined values of $V_{n2,2}$ and $Q_{1,1}$ must be tuned properly.

*1.1.1.2 Transient response analysis*

Besides analyzing the occurrence of the robot being able to reject disturbances, the transient response is also analyzed. The robot's transient response when subject to disturbances from the pendulum is shown in **Fig. 14** and **Fig. 15**. At first, the robot moves in the direction of the disturbance. This is shown by the CoM angle moving from an initial value of 3.416° to -22.21° because initially, the control effort from the ankle pitch servos was not able to resist the disturbance. As the control effort is being applied to the ankle pitch servos, the CoM angle begins to move to the equilibrium. This causes a change in the direction of the estimated angular velocity as it starts to move in the direction opposite to the CoM angle and then causing the control effort to decrease. However, this decreasing control effort causes the CoM angle to increase. When the CoM angle increases, the control effort increases to reduce the CoM angle. This is what causes the oscillations. Nonetheless, the controller can return the CoM angle to its equilibrium point with a final value of 2.391°.

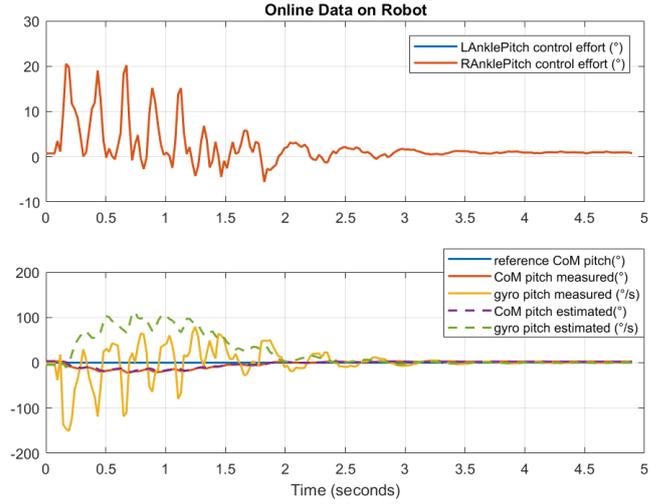

**Fig. 14.** Transient response of all states

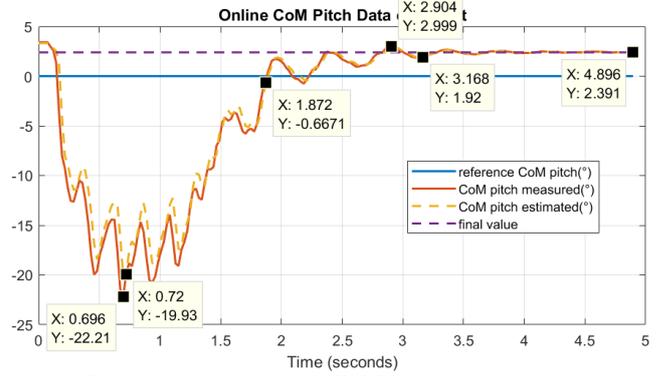

**Fig. 15.** Transient response of the CoM angle state

From the transient response in **Fig. 15**, the experimental results can be calculated then compared with the expected specifications shown in **Table 7**. These specifications are based on previous research [14].

Rise time is the time required for the output response to rise from 10% to 90% of its final value [31]. From **Fig. 15**, the time when the output response is at 10% and 90% of its final value can be used to calculate the rise time:

$$t_r = t_{90\%} - t_{10\%} \quad (18)$$
$$= 1.872 - 0.72$$
$$= 1.152 \text{ seconds.}$$

Settling time is the time required for the output response to reach and stay within a range of about 2% of its final value [31]. From **Fig. 15**, the time when the output response is at its initial value and 2% of its final value can be used to calculate the settling time:

$$t_s = t_{2\%final} - t_{initial} \quad (19)$$
$$= 3.168 - 0.696$$
$$= 2.472 \text{ seconds.}$$

The maximum overshoot can be measured as the absolute value of the maximum value that exceeds the final value:

$$OS_{max} = |y_{max} - y_{final}| \quad (20)$$
$$= |3.046 - 2.391|$$
$$= 0.655 \text{ degrees.}$$

The CoM angle robustness to disturbance can be measured as the absolute value of the difference between its minimum and maximum values:

$$\Delta y = |y_{max} - y_{min}| \quad (21)$$
$$= 2.391 - (-22.21)$$
$$= 24.601 \text{ degrees.}$$

**Table 7.** Transient response specification results

| No. | Specification | Unit | Expected | Result |
|---|---|---|---|---|
| 1 | Rise time | seconds | Less than or equal to 1 | 1.152 |
| 2 | Settling time | seconds | Less than or equal to 2 | 2.472 |
| 3 | Maximum overshoot of CoM angle | degrees | Maximum 7.5 | 0.655 |
| 4 | Steady-state error of CoM angle | degrees | Maximum 2.5 | 2.391 |
| 5 | CoM angle robustness to disturbance | degrees | Minimum 10 (±5) | 24.601 |

Even though the rise time and settling time specifications in **Table 7** are not satisfied, the difference between the expected and results are negligible. This may be caused by the suboptimal trial-error tuning process. Nonetheless, the robot can still maintain its stability when subject to disturbances from a swinging pendulum.

Even though there are oscillations in the transient response, this controller configuration is still better than using raw IMU data without an observer or using only the CoM angle for feedback. If we only use raw IMU data, with the same $K$ gain, the robot's oscillation will gradually increase and the robot will fall. This is caused by the noisy gyro sensor. If we only use the CoM angle without angular velocity, the robot will never stop oscillating even though the CoM angle is small. This is similar to the undamped response of a P-controller.

*8.1.3    Fuzzy logic control*

To reject disturbances that the model-based controller couldn't handle, a fuzzy logic approach is implemented.

*1.1.1.3  Tuning of membership functions and rules*

The membership functions explained in chapter 6.2.2 are declared and tuned based on the robot's response when subject to disturbances from chapter 8.1.2.

(1) Angle input membership function

The asymmetrical trapezoidal shape of the angle input membership function shown in **Fig. 16** is to compensate for the robot's slightly leaning forward posture.

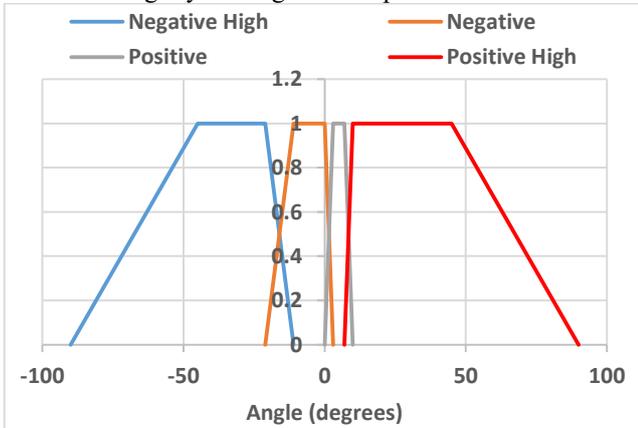

**Fig. 16.** Angle input membership function graphical form
This membership function can also be represented in a tabular form shown in **Table 8** based on the membership function structure explained in **Fig. 8**.

**Table 8.** Angle input membership function tabular form

| Membership | $b_1$ | $u_1$ | $u_2$ | $b_2$ |
|---|---|---|---|---|
| Negative High | -90 | -45 | -21 | -11 |
| Negative | -21 | -11 | 0 | 3 |
| Positive | 0 | 3 | 7 | 10 |
| Positive High | 7 | 10 | 45 | 90 |

(2) Angular velocity input membership function

The angular velocity input membership function shown in **Fig. 17** is based on the tendency of the angular velocity data to oscillate at a larger range ($|\dot{\theta}| \geq 3$ rad/s) represented by the Negative and Positive memberships. On the other hand, the Average membership represents the angular velocity when its oscillation is at its minimum ($|\dot{\theta}| \leq 0.5$ rad/s).

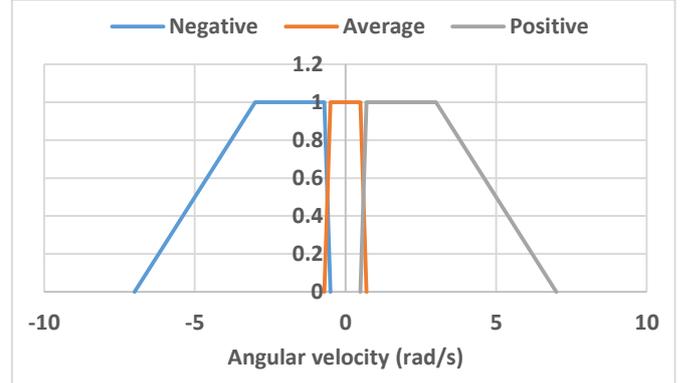

**Fig. 17.** Angular velocity input membership function graphical form

This membership function can also be represented in a tabular form shown in **Table 9**.

**Table 9.** Angular velocity input membership function tabular form

| Membership | $b_1$ | $u_1$ | $u_2$ | $b_2$ |
|---|---|---|---|---|
| Negative | -7 | -3 | -0.7 | -0.5 |
| Average | -0.7 | -0.5 | 0.5 | 0.7 |
| Positive | 0.5 | 0.7 | 3 | 7 |

(3) Angle gain output membership function

The output membership function shown in **Fig. 18** is based on the LQR angle gain $K_\theta$ on various $Q_{1,1}$ state cost matrix angle weights.

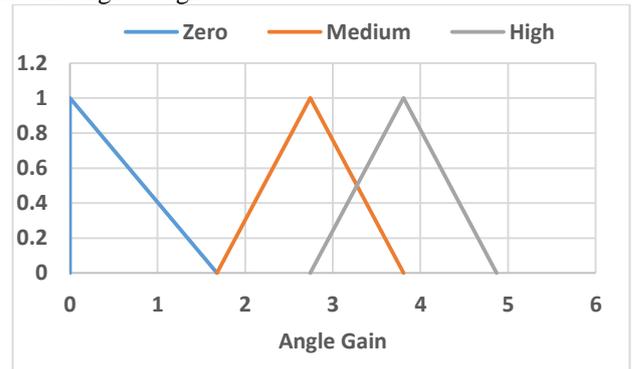

**Fig. 18.** Angle gain output membership function graphical form
The Zero $Q_{1,1}$ weight is to improve the robot's response to settle in its equilibrium point. The Medium $Q_{1,1}$ weight is based on the results from chapter 1.1.1.1. The High $Q_{1,1}$ weight is tuned such that the control effort increases at higher angles but not too high to cause overshoot. The chosen $Q_{1,1}$ values and their respective LQR angle gain are shown in **Table 10**.

**Table 10.** $Q_{1,1}$ values and their respective LQR angle gain

| Membership | $Q_{1,1}$ | $K_\theta$ |
|---|---|---|
| Zero (Z) | 0 | 0 |
| Medium (M) | 40 | 2.743 |
| High (H) | 75 | 3.806 |

The output membership function in **Fig. 18** can also be represented in a tabular form shown in **Table 11**.

**Table 11.** Angle gain output membership function tabular form

| Membership | $b_1$ | $u_1$ | $u_2$ | $b_2$ |
|---|---|---|---|---|
| Zero (Z) | 0 | 0 | 0 | 1.68 |
| Medium (M) | 1.68 | 2.743 | 2.743 | 3.806 |
| High (H) | 2.743 | 3.806 | 3.806 | 4.869 |

(4) Angular velocity gain output membership function

Similar to section (3), the output membership shown in **Fig. 19** is based on the LQR angular velocity gain $K_{\dot\theta}$ on various $Q_{1,1}$ weights.

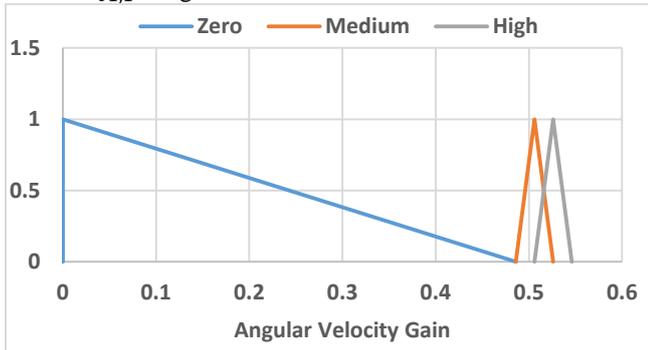

**Fig. 19.** Angular velocity gain output membership function graphical form

The values of these gains are shown in **Table 12**.

**Table 12.** $Q_{1,1}$ values and their respective LQR angular velocity gain

| Membership | $Q_{1,1}$ | $K_{\dot\theta}$ |
|---|---|---|
| Zero (Z) | 0 | 0 |
| Medium (M) | 40 | 0.506 |
| High (H) | 75 | 0.526 |

A tabular representation of **Fig. 19** is shown in **Table 13**.

**Table 13.** Angular velocity gain output membership function tabular form

| Membership | $b_1$ | $u_1$ | $u_2$ | $b_2$ |
|---|---|---|---|---|
| Zero (Z) | 0 | 0 | 0 | 0.486 |
| Medium (M) | 0.486 | 0.506 | 0.506 | 0.526 |
| High (H) | 0.506 | 0.526 | 0.526 | 0.546 |

Two fuzzy rules must also be declared because there are two fuzzy systems (angle gain $K_\theta$ and angular velocity gain $K_{\dot\theta}$). The rules shown in **Table 14** and **Table 15** are tuned based on the response from chapter 8.1.2 when the ankle strategy is applied without fuzzy configuration.

**Table 14.** Angle gain rule

| Angular velocity \ Angle | Negative (N) | Average (A) | Positive (P) |
|---|---|---|---|
| Negative High (NH) | H | H | H |
| Negative (N) | M | M | M |
| Positive (P) | Z | Z | Z |
| Positive High (PH) | H | H | H |

**Table 15.** Angular velocity gain rule

| Angular velocity \ Angle | Negative (N) | Average (A) | Positive (P) |
|---|---|---|---|
| Negative High (NH) | H | H | H |
| Negative (N) | M | M | M |
| Positive (P) | Z | Z | Z |
| Positive High (PH) | M | M | M |

where the output fuzzy sets Z=1, M=2, and H=3 when the rules are applied in **Algorithm 2**.

*1.1.1.4 Results analysis*

The ankle strategy experimental result using various configurations is shown in **Table 16**. This is repeated five times for each angle. Out of the five trials, we counted the number of times the robot is still able to stand stably. For example, the robot is able to stand three times and falls two times, so the probability of the robot standing stably for that amplitude angle is 3/5. For the next iterations, the swinging angle is increased by 1 degree and several configuration variations. This is done to ensure that the result isn't just pure luck. The black patches indicate for that amplitude angle the robot is guaranteed to fall.

Based on the results shown in **Table 16**, the fuzzy-extended controller's ability to change the state feedback gain value can increase the system's robustness compared to model-based control without fuzzy logic. The reason is that the fuzzy logic can handle nonlinearities in the system when the CoM angle is large. When the CoM angle is too large, the model acquired from system identification is obsolete and the foot contact point with the grass is already too small. Because of that, a knowledge-based control approach is required which is a fuzzy-extended controller that recognizes a significantly larger control effort is required for a larger CoM angle.

A more generalized representation of the disturbance acting on the robot can be represented by the amount of kinetic energy when the pendulum hits the robot's torso. With prior knowledge of the pendulum's length $L$, the amplitude angle $\theta$, and the pendulum's mass $m$, the kinetic energy can be calculated using conservation of mechanical energy equation $KE_{after} = mgL(1 - \cos\theta)$.

By substituting the total mass of the pendulum $m = 0.76$ kg dan total length of the pendulum $L = 1$ meter, **Table 16** can be converted to kinetic energy representation as shown in **Table 17**.

**Table 16.** Ankle strategy testing results

| No. | Disturbance origin | Configuration | Pendulum amplitude angle (degrees) | | | | | |
|---|---|---|---|---|---|---|---|---|
| | | | 24 | 25 | 26 | 27 | 28 | 29 |
| 1 | In front of the robot | No feedback | | 5/5 | | | | |
| 2 | | Feedback without fuzzy | | | 4/5 | | | |
| 3 | | Feedback with fuzzy | | | 2/5 | 4/5 | | |
| 4 | Behind the robot | No feedback | | | 3/5 | | | |
| 5 | | Feedback without fuzzy | 3/5 | | | | | |
| 6 | | Feedback with fuzzy | | | | | 4/5 | |

**Table 17.** Kinetic energy representation of ankle strategy testing results

| No. | Disturbance origin | Configuration | Kinetic energy (joules) | | | | | |
|---|---|---|---|---|---|---|---|---|
| | | | 0.645 | 0.699 | 0.755 | 0.813 | 0.873 | 0.935 |
| 1 | In front of the robot | No feedback | | 5/5 | | | | |
| 2 | | Feedback without fuzzy | | | 4/5 | | | |
| 3 | | Feedback with fuzzy | | | 2/5 | 4/5 | | |
| 4 | Behind the robot | No feedback | | | 3/5 | | | |
| 5 | | Feedback without fuzzy | 3/5 | | | | | |
| 6 | | Feedback with fuzzy | | | | | 4/5 | |

**Table 18.** Stepping strategy testing results

| No. | 1. | 2. | 3. | 4. | 5. |
|---|---|---|---|---|---|
| **Configuration** | No feedback | Ankle strategy without fuzzy | Ankle strategy with fuzzy | Ankle strategy + Capture Point without fuzzy | Ankle strategy + Capture Point with fuzzy |
| **Average maximum force (newtons)** | 1.94 | 3.5 | 3.12 | 3.64 | 4.4 |

*8.2   Stepping strategy*

The stepping strategy is evaluated by testing the robot's robustness when subject to steady force. This is achieved by commanding the robot to walk forward while its hip is attached to a spring scale using a string and is fixed on a pillar as shown in **Fig. 20**. The purpose of this experiment is to know the maximum force the robot can pull before falling. In this experiment, we applied the QuinticWalk gait [22] with a walking amplitude of 0.03 meters and a walking frequency of 1.6 Hz.

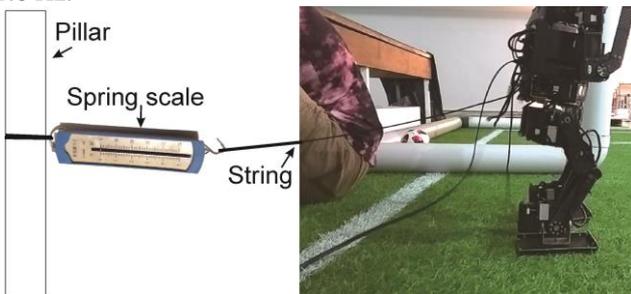

**Fig. 20.** Stepping strategy experimental setup

For every configuration variation, the experiment is conducted five times. From these five measurements, the average is calculated. The stepping strategy effect on the robot was tested and results are shown in **Table 18**.

Based on **Table 18** numbers 4 and 5, the ankle strategy and capture point with fuzzy produces a higher average maximum force compared to the ankle strategy and capture point without fuzzy. The configuration in **Table 18** number 5 can withstand the highest average maximum force compared to other configurations with twice the value of the no feedback configuration (**Table 18** number 1).

If we analyze the effect of capture point from **Table 18** numbers 4 and 5 compared to without capture point from **Table 18** numbers 2 and 3, the capture point configuration increases the maximum force that the robot can withstand. The reason is that when the robot is walking forward, the spring scale pulls the robot backward causing the robot to lean backward. The capture point algorithm detects this disturbance and calculates the walking amplitude in the $x$-axis that the robot must place its foot onto so it wouldn't fall.

If we analyze the effect of the fuzzy algorithm on the stepping strategy (**Table 18** number 5) compared to without fuzzy (**Table 18** number 4), the fuzzy algorithm can increase the maximum force the robot can withstand. The fuzzy algorithm can automatically change the state feedback gain according to the CoM angle and angular velocity state of the robot. This algorithm can also handle the system's nonlinearity that is a significantly larger control effort is required to stabilize the robot when the CoM angle is larger.

9 **Conclusion**

The implemented controller in this paper can maintain a position-controlled robot's stability around the pitch or $y$-axis when subject to disturbances on a standing pose or even walking on synthetic grass. The robot's stability is maintained by controlling the CoM angle to approximate the equilibrium or zero.

The model acquired from system identification is applied as the basis for the design of the controller and Kalman Filter.

Using only the LQR ankle strategy, the transient response specifications are satisfied except for the rise time and settling time with negligible difference. The fuzzy-extended LQR ankle strategy is proven to improve the robot's robustness to disturbances when tested on a standing pose and walking. The combined fuzzy-extended LQR ankle strategy and the capture point stepping strategy can withstand the highest average maximum force compared to other configurations when tested during walking.

## 10 Future work

To achieve a better transient response, optimization and learning algorithms can be applied to the fuzzy membership functions. This controller is also only capable of rejecting disturbances from the pitch or $y$-axis. Thus, a roll or $x$-axis controller is necessary. To increase its robustness, hip and/or arm strategy can be utilized.

## Appendix A : Individual area and centroid calculation

If the section's shape is a triangle as shown in **Fig. 21**,

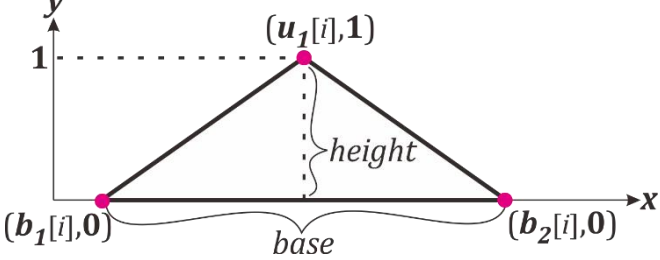

**Fig. 21.** Triangle membership function

the area can be calculated as:
$$Area = \frac{|base| \cdot |height|}{2} \quad (A.1)$$
$$= \frac{|b_2[i] - b_1[i]| \cdot |1 - 0|}{2}$$
$$= \frac{|b_2[i] - b_1[i]|}{2}$$

while the centroid can be calculated as:
$$CoM_x = \frac{b_1[i] + u_1[i] + b_2[i]}{3}. \quad (A.2)$$

If the section's shape is a trapezoid as shown in **Fig. 22**,

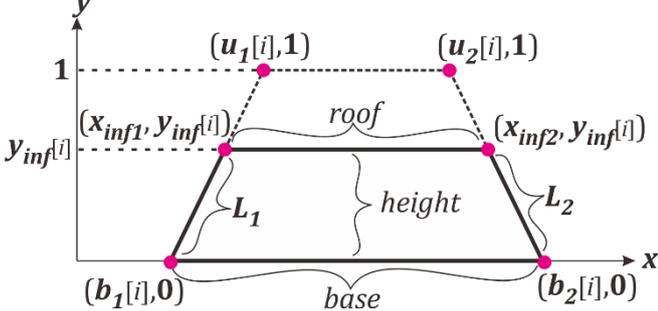

**Fig. 22.** Trapezoid membership function

the length of the trapezoid's legs $L_1$ and $L_2$ can be calculated by first locating $x_{inf1}$ and $x_{inf2}$, which are the $x$-axis values that intersect the trapezoid's legs at $y = y_{inf}[i]$:

$$x_{inf1} = \frac{x_2 - x_1}{y_2 - y_1} \cdot (y - y_1) + x_1 \quad (A.3)$$
$$= \frac{u_1[i] - b_1[i]}{1 - 0} \cdot (y_{inf}[i] - 0) + b_1[i]$$
$$= (u_1[i] - b_1[i]) \cdot y_{inf}[i] + b_1[i]$$

$$x_{inf2} = \frac{x_2 - x_1}{y_2 - y_1} \cdot (y - y_1) + x_1 \quad (A.4)$$
$$= \frac{b_2[i] - u_2[i]}{0 - 1} \cdot (y_{inf}[i] - 1) + u_2[i]$$
$$= (u_2[i] - b_2[i]) \cdot (y_{inf}[i] - 1) + u_2[i].$$

After acquiring $x_{inf1}$ and $x_{inf2}$, the length of the trapezoid's legs $L_1$ and $L_2$ can be calculated as:

$$|L_1| = \sqrt{(x_2 - x_1)^2 + (y_2 - y_1)^2} \quad (A.5)$$
$$= \sqrt{(x_{inf1} - b_1[i])^2 + (y_{inf}[i] - 0)^2}$$
$$= \sqrt{(x_{inf1} - b_1[i])^2 + (y_{inf}[i])^2}$$

$$|L_2| = \sqrt{(x_2 - x_1)^2 + (y_2 - y_1)^2} \quad (A.6)$$
$$= \sqrt{(b_2[i] - x_{inf2})^2 + (0 - y_{inf}[i])^2}$$
$$= \sqrt{(b_2[i] - x_{inf2})^2 + (y_{inf}[i])^2}.$$

The length of the trapezoid's $roof$, $base$, and $height$ can be calculated as:
$$|roof| = |x_{inf2} - x_{inf1}| \quad (A.7)$$
$$|base| = |b_2[i] - b_1[i]|. \quad (A.8)$$

Then the area and centroid of the trapezoid can be calculated by substituting the values acquired from Equations (A.3)-(A.8) into Equations (A.9) and (A.10) respectively:

$$Area = \frac{(|base| + |roof|) \cdot |height|}{2} \quad (A.9)$$

$$CoM_x \quad (A.10)$$
$$= b_1[i] + \frac{|base|}{2}$$
$$+ \frac{(2 \cdot |roof| + |base|) \cdot (|L_1|^2 - |L_2|^2)}{6 \cdot (|base|^2 - |roof|^2)}$$

## Appendix B : Intersection area and centroid calculation

The intersection between two memberships can be categorized into three types of conditions based on its $y_{inf}$ inference value in the $y$-axis as shown in **Fig. 23** until **Fig. 25**. **Fig. 23** shows the condition when $y_{inf}[j]$ and $y_{inf}[j+1]$ are higher than the intersection's peak. **Fig. 24** shows the condition when $y_{inf}[j]$ or $y_{inf}[j+1]$ is equal to the intersection's peak. **Fig. 25** shows the condition when $y_{inf}[j]$ or $y_{inf}[j+1]$ is lower than the intersections' peak.

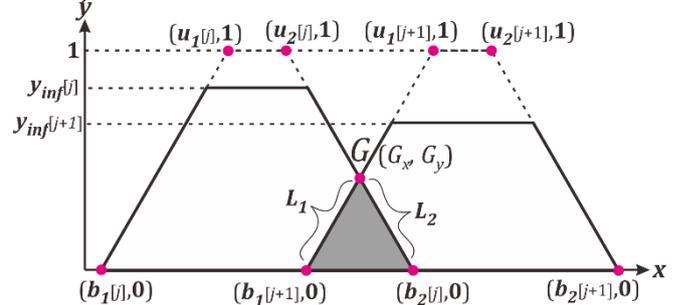

**Fig. 23.** Intersection model 1

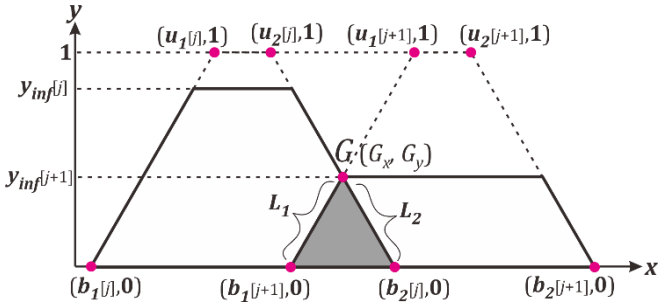
**Fig. 24.** Intersection model 2

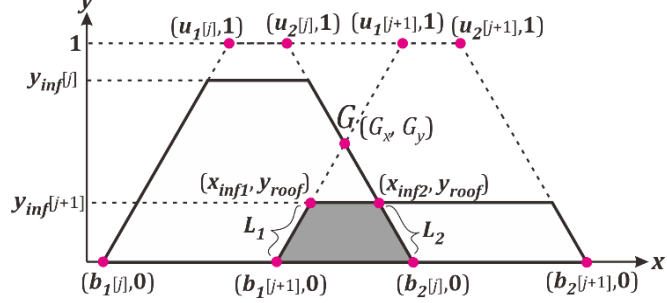
**Fig. 25.** Intersection model 3

First, we must locate the coordinate of the intersection triangle's peak that is represented by point $G$ in **Fig. 23** until **Fig. 25**. To accomplish this, the equations that describe lines $L_1$ and $L_2$ must be obtained first. The gradient of lines $L_1$ and $L_2$ respectively ($m_1$ and $m_2$) can be calculated as:

$$m_1 = \frac{y_2 - y_1}{x_2 - x_1} \quad \text{(B.11)}$$
$$= \frac{1 - 0}{u_1[j+1] - b_1[j+1]}$$
$$= \frac{1}{u_1[j+1] - b_1[j+1]}$$
$$m_2 = \frac{y_2 - y_1}{x_2 - x_1} \quad \text{(B.12)}$$
$$= \frac{0 - 1}{b_2[j] - u_2[j]}$$
$$= -\frac{1}{b_2[j] - u_2[j]}.$$

After acquiring the gradient of both lines, the value of constant $c$ of both memberships can be calculated as:

$$y_1 = (m_1 \cdot x_1) + c_1 \quad \text{(B.13)}$$
$$c_1 = y_1 - (m_1 \cdot x_1)$$
$$= 1 - (m_1 \cdot u_1[j+1])$$
$$y_2 = (m_2 \cdot x_2) + c_2 \quad \text{(B.14)}$$
$$c_2 = y_2 - (m_2 \cdot x_2)$$
$$= 0 - (m_2 \cdot b_1[j])$$
$$= -m_2 \cdot b_1[j].$$

After acquiring the gradients ($m_1$ and $m_2$) and the constants ($c_1$ and $c_2$), the coordinates of point $G$ can then be calculated using the concept of two lines intersecting on a point:

$$y_1 = y_2 \quad \text{(B.15)}$$
$$m_1 \cdot G_x + c_1 = m_2 \cdot G_x + c_2$$
$$(m_1 - m_2) \cdot G_x = c_2 - c_1$$
$$G_x = \frac{c_2 - c_1}{m_1 - m_2}$$

$$G_y = \frac{0 - 1}{b_2[j] - u_2[j]} \cdot (G_x - u_2[j]) + 1 \quad \text{(B.16)}$$
$$= 1 - \frac{G_x - u_2[j]}{b_2[j] - u_2[j]}.$$

For intersection models 1 and 2 shown in **Fig. 23** and **Fig. 24**, the area and centroid of this triangle intersection can be calculated as:

$$Area = \frac{|base| \cdot |height|}{2} \quad \text{(B.17)}$$
$$= \frac{|b_2[j] - b_1[j+1]| \cdot |G_y - 0|}{2}$$
$$= \frac{|b_2[j] - b_1[j+1]| \cdot |G_y|}{2}$$
$$CoM_x = \frac{b_1[j+1] + G_x + b_2[j]}{3}. \quad \text{(B.18)}$$

For intersection model 3 shown in **Fig. 25**, the shape of the intersection is a trapezoid with its height substituted with the lower value between $y_{inf}[j]$ or $y_{inf}[j+1]$:

$$y_{roof} = \min(y_{inf}[j], y_{inf}[j+1]). \quad \text{(B.19)}$$

In this case, $y_{roof} = y_{inf}[j+1]$. The coordinates of $x_{inf1}$ and $x_{inf2}$ of the trapezoid can be calculated by modifying Equations (A.3) and (A.4):

$$x_{inf1} = \frac{x_2 - x_1}{y_2 - y_1} \cdot (y - y_1) + x_1 \quad \text{(B.20)}$$
$$= \frac{u_1[j+1] - b_1[j+1]}{1 - 0} \cdot (y_{roof} - 0) + b_1[j+1]$$
$$= (u_1[j+1] - b_1[j+1]) \cdot y_{roof} + b_1[j+1]$$
$$x_{inf2} = \frac{x_2 - x_1}{y_2 - y_1} \cdot (y - y_1) + x_1 \quad \text{(B.21)}$$
$$= \frac{b_2[j] - u_2[j]}{0 - 1} \cdot (y_{roof} - 1) + u_2[j]$$
$$= (u_2[j] - b_2[j]) \cdot (y_{roof} - 1) + u_2[j].$$

After that, the length of the trapezoid's legs can be calculated by modifying Equations (A.5) and (A.6):

$$|L_1| = \sqrt{(x_2 - x_1)^2 + (y_2 - y_1)^2} \quad \text{(B.22)}$$
$$= \sqrt{(x_{inf1} - b_1[j+1])^2 + (y_{roof} - 0)^2}$$
$$= \sqrt{(x_{inf1} - b_1[j+1])^2 + (y_{roof})^2}$$
$$|L_2| = \sqrt{(x_2 - x_1)^2 + (y_2 - y_1)^2} \quad \text{(B.23)}$$
$$= \sqrt{(b_2[j] - x_{inf2})^2 + (0 - y_{roof})^2}$$
$$= \sqrt{(b_2[j] - x_{inf2})^2 + (y_{roof})^2}.$$

The trapezoid's area and centroid can then be calculated using Equations (A.9) and (A.10) respectively.

## 12 Declaration of competing interest

This work is part of the research done by the first and second authors who are students of the Department of Electrical and Information Engineering, Universitas Gadjah Mada.